\documentclass[sigconf,nonacm]{acmart}

\renewcommand\footnotetextcopyrightpermission[1]{}

\AtBeginDocument{%
  }

\setcopyright{none}
\copyrightyear{2026}
\acmYear{2026}
\acmDOI{}
\acmConference[Conference '26]{Conference}{2026}{Location}
\acmISBN{}

\usepackage{algorithm}
\usepackage{algorithmic}
\usepackage{eso-pic}
\usepackage{graphicx}
\usepackage[table]{xcolor}
\usepackage{tikz}
\usepackage{fontawesome5}
\usepackage{bbding}

\newcommand{\corresponding}{%
  \textsuperscript{\scriptsize\Envelope}%
}

\makeatletter
\newcommand{\correspondingnote}{%
  \g@addto@macro\@authornotes{%
    \begingroup
    \renewcommand{\thefootnote}{\Envelope}%
    \stepcounter{footnote}%
    \footnotetext{Corresponding author.}%
    \endgroup
  }%
}
\makeatother

\definecolor{NavyBlue}{RGB}{0,38,84}
\definecolor{TealGreen}{RGB}{0,128,128}

\definecolor{BrainB}{RGB}{0,38,84}
\definecolor{Brainr}{RGB}{0,49,88}
\definecolor{Braina}{RGB}{0,60,92}
\definecolor{Braini}{RGB}{0,71,96}
\definecolor{Brainn}{RGB}{0,82,101}
\definecolor{BrainW}{RGB}{0,94,106}
\definecolor{BrainA}{RGB}{0,105,111}
\definecolor{BrainM}{RGB}{0,117,119}

\newcommand{\GradientBrainWAM}{%
  {\textcolor{BrainB}{B}%
   \textcolor{Brainr}{r}%
   \textcolor{Braina}{a}%
   \textcolor{Braini}{i}%
   \textcolor{Brainn}{n}%
   \textcolor{BrainW}{W}%
   \textcolor{BrainA}{A}%
   \textcolor{BrainM}{M}}%
}

\begin{document}

\title{\GradientBrainWAM: Action-Space Coordination of Semantic Priors and Predictive Dynamics for Autonomous Driving}

\author{Bing Zhan}
\authornote{Equal contribution.}
\affiliation{%
  \institution{NLPR, Institute of Automation, Chinese Academy of Sciences (CASIA)}
  \city{Beijing}
  \country{China}
}

\author{Shuyao Shang}
\authornotemark[1]
\affiliation{%
  \institution{NLPR, Institute of Automation, Chinese Academy of Sciences (CASIA)}
  \city{Beijing}
  \country{China}
}

\author{Shuo Lu}
\affiliation{%
  \institution{NLPR, Institute of Automation, Chinese Academy of Sciences (CASIA)}
  \city{Beijing}
  \country{China}
}

\author{Yuan Xu}
\affiliation{%
  \institution{NLPR, Institute of Automation, Chinese Academy of Sciences (CASIA)}
  \city{Beijing}
  \country{China}
}

\author{Zhao Wang}
\affiliation{%
  \institution{Li Auto Inc.}
  \city{Beijing}
  \country{China}
}

\author{Yida Wang}
\affiliation{%
  \institution{Li Auto Inc.}
  \city{Beijing}
  \country{China}
}

\author{Xueyang Zhang}
\affiliation{%
  \institution{Li Auto Inc.}
  \city{Beijing}
  \country{China}
}

\author{Kun Zhan}
\affiliation{%
  \institution{Li Auto Inc.}
  \city{Beijing}
  \country{China}
}

\author{Jiahao Gu\corresponding}
\authornote{Project leader.}
\affiliation{%
  \institution{Li Auto Inc.}
  \city{Beijing}
  \country{China}
}

\correspondingnote

\renewcommand{\shortauthors}{Zhan et al.}

\begin{abstract}
Autonomous driving requires planning under both semantic constraints and predictive dynamics. Existing end-to-end driving approaches, however, typically emphasize only one side of this requirement: Vision-Language-Action (VLA) models exploit VLM priors for semantic reasoning, while World Action Models (WAMs) provide future-aware prediction through generative world modeling. This naturally motivates a unified planner that can leverage both semantic priors and predictive dynamics. However, we find that a naive combination through joint token-level attention suffers from an attention-allocation mismatch, where semantic shortcuts dominate the shared attention space and suppress predictive dynamics. Inspired by neuroscience evidence that complex behavior arises from coordination among functionally specialized systems, we propose BrainWAM, a structured action-space coordination framework that converts semantic reasoning and predictive world modeling into two specialized action-oriented pathways, and aligns them at the level of compact action representations. We further introduce an asynchronous rectified-flow inference strategy with decoupled video and action denoising, which shortens inference latency while preserving planning-relevant predictive context. BrainWAM reaches state-of-the-art performance on both NAVSIM v1 ($89.5$ PDMS) and NAVSIM v2 ($89.6$ EPDMS), consistently outperforming VLA-only or WAM-only methods, highlighting BrainWAM as a practical and promising direction for autonomous driving systems.
\end{abstract}


\maketitle
\begin{figure}[!t]
    \centering
    \includegraphics[width=\columnwidth]{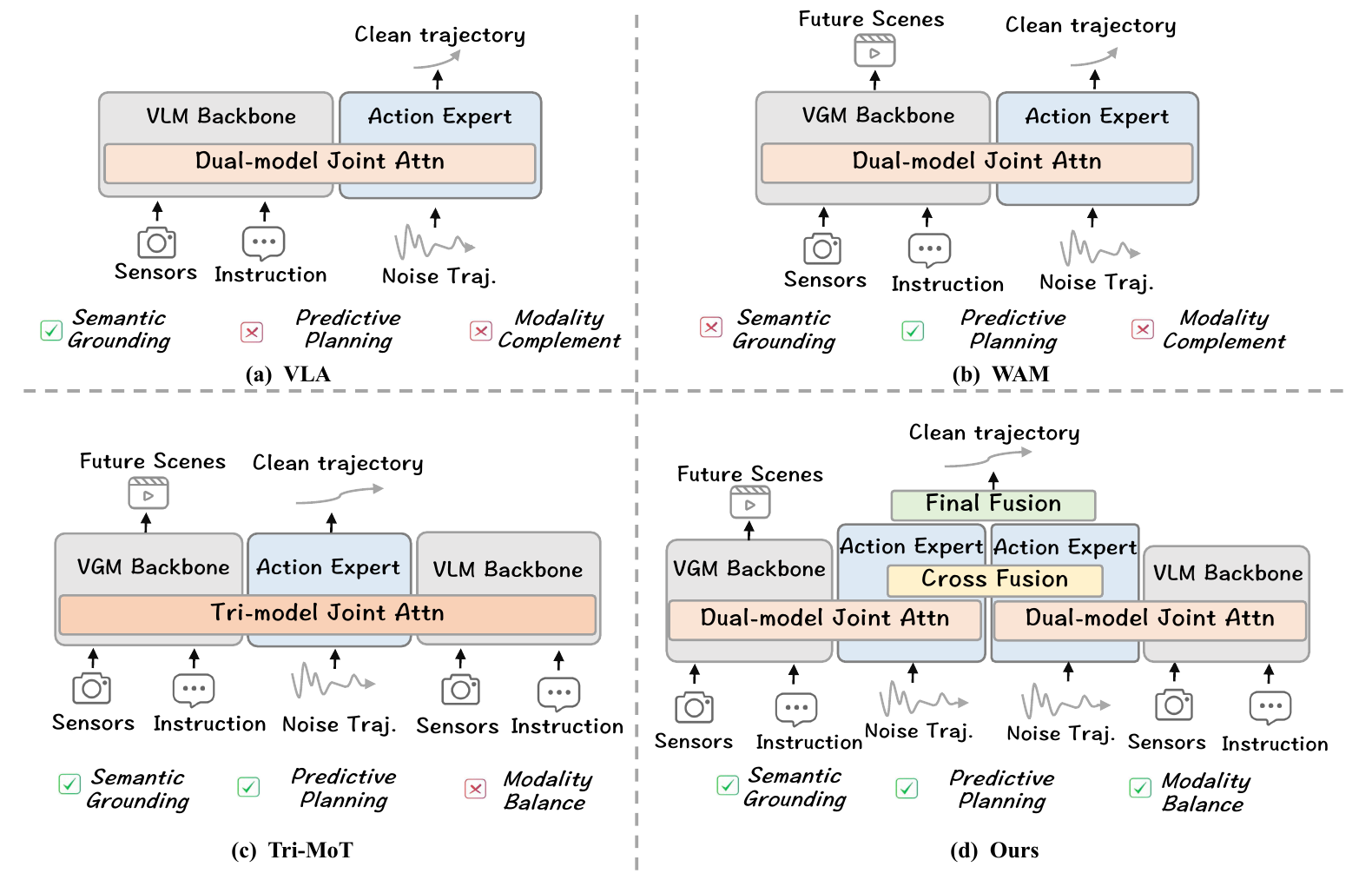}
    \caption{
    \textbf{Comparison of different paradigms in autonomous driving.}
    (a) VLA leverages vision-language priors for task-aware semantic grounding but lacks explicit predictive planning.
    (b) WAM captures future scene evolution but has limited semantic grounding.
    (c) Tri-MoT jointly fuses VLM, VGM, and action tokens in a shared raw-token space, which may cause attention interference.
    (d) Our method separates semantic and predictive pathways and coordinates them in the action space. VGM: Video Generation Model.
    }
    \label{fig:teaser}
\end{figure}

\section{Introduction}
\label{sec:intro}

Autonomous driving requires planning under two tightly coupled forms of evidence: semantic constraints and predictive dynamics.
Vision-Language-Action (VLA) models leverage the world-knowledge priors of Vision-Language Models (VLMs), making them effective at grounding observations in traffic rules, route instructions, scene semantics, and high-level driving intent.
World Action Models (WAMs), inspired by recent progress in action-conditioned world modeling~\citep{uwm,videovla,motus}, instead learn how actions and future states evolve together, providing predictive context for motion trends, interaction outcomes, and physical feasibility.
These strengths are naturally complementary: VLA models provide task-aware semantic and decision priors but usually lack explicit modeling of future scene evolution, whereas WAMs provide future-aware dynamics and physical priors but are less reliable at rule-aware and intent-driven reasoning.
This raises our central question: \emph{how can VLA and WAM be effectively combined to unleash the complementary potential of semantic reasoning and predictive modeling?}

\begin{figure}[htbp]
    \centering
    \includegraphics[width=\columnwidth]{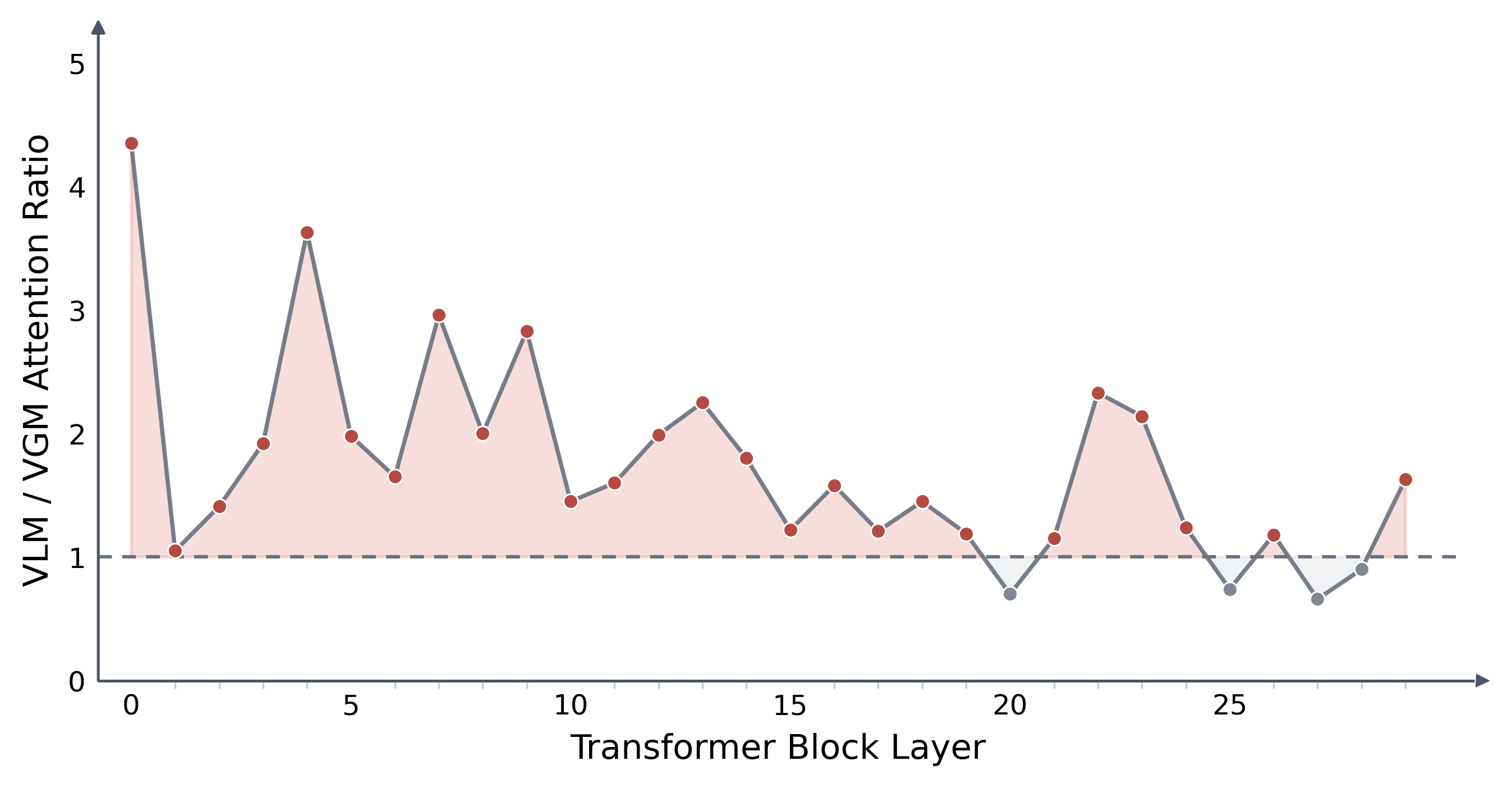}
    \caption{\textbf{Attention allocation in Tri-MoT.}
    We compare attention ratios of action tokens to VLM and VGM tokens across layers.
    Action tokens attend more strongly to VLM tokens than to VGM tokens
    across most Transformer layers,
    revealing semantic dominance in joint representation space.}
    \label{fig:trimot_attn}
\end{figure}

A common direct design is Tri-modal Joint Attention (Tri-MoT), which places VLM tokens, Video Generative Model (VGM) tokens, and action tokens into one shared attention space.
However, we find that this raw-token fusion can even underperform WAM alone.
To diagnose this issue, we visualize how action tokens attend to VLM and VGM tokens.
As shown in Fig.~\ref{fig:trimot_attn}, action tokens attend more strongly to semantic-level VLM tokens than to pixel-level VGM tokens across most Transformer layers, especially in shallow layers.
This asymmetry follows the modality competition observed in joint multimodal training~\citep{gradient_blending,modality_competition,ogm_ge,unimodal_feature_learning}, where the modality that is easier to learn dominates optimization and suppresses the other (see Appendix A).
Here the clean and semantically compact VLM tokens are easier to learn, while the VGM tokens are still being denoised and provide lower-signal features, so action tokens take the VLM shortcut and underuse the predictive video tokens.
As a result, directly mixing high-dimensional heterogeneous tokens induces an \emph{attention-allocation mismatch}: semantic signals dominate the shared interaction space and weaken the predictive dynamics needed for planning.

To address this challenge, we draw inspiration from neuroscience: complex behavior emerges not from homogenizing all signals into one undifferentiated representation, but from coordination among functionally specialized systems.
The left hemisphere is often associated with language, symbolic, and sequential processing, while the right hemisphere plays an important role in visuospatial and holistic understanding; the two hemispheres exchange information through the corpus callosum, and motor intent is further coordinated and refined by the cerebellum~\citep{gazzaniga2005,kanwisher2010,wolpert1998,buckner2013,bostan2018}.
This organization suggests a computational principle for VLA-WAM integration: semantic reasoning and predictive world modeling should first develop specialized, behavior-relevant action representations, and then coordinate through compact action-level communication.

Motivated by this principle, we propose \textbf{BrainWAM}, a brain-inspired action-space coordination framework for autonomous driving.
BrainWAM structures semantic reasoning and predictive world modeling into two complementary action-oriented pathways: a left-hemisphere pathway distills traffic-scene semantics, route instructions, and rule-aware decision priors from VLM, while a right-hemisphere pathway distills spatiotemporal dynamics, physical consistency, and future-interaction cues from VGM.
The two pathways communicate bidirectionally over compact action tokens through a corpus-callosum-inspired Callosal Action Bridge (CAB), and a cerebellum-inspired Cerebellar Intent Fusion (CIF) module coordinates the refined action intents and decodes them into an executable trajectory.
Experiments on NAVSIM v1 and v2 demonstrate the effectiveness of this design: BrainWAM consistently outperforms VLA-only, WAM-only, and raw-token fusion baselines, and achieves state-of-the-art performance over existing end-to-end driving, VLA-based, and world-model-based methods.

We summarize our contributions as follows:
\begin{itemize}

\item We propose BrainWAM, an action-level coordination framework that combines VLM-based semantic reasoning with WAM-based predictive world modeling. Inspired by brain functional specialization, BrainWAM converts instruction-aware semantic constraints and future-dynamics priors into complementary action representations, and coordinates them in a unified action space.

\item We identify an \emph{attention-allocation mismatch} in Tri-MoT: its action tokens attend disproportionately to semantic tokens in most layers, causing raw-token fusion to underperform WAM-only planning.

\item BrainWAM achieves 89.5 PDMS on NAVSIM v1 and 89.6 EPDMS on NAVSIM v2, outperforming strong end-to-end driving, VLA-based, and world-model-based methods. These results show BrainWAM's feasibility and potential for autonomous driving systems.

\end{itemize}

\section{Related Work}
\label{sec:related}

\begin{figure*}[!t]
    \centering
    \includegraphics[width=\textwidth,height=0.32\textheight,keepaspectratio]{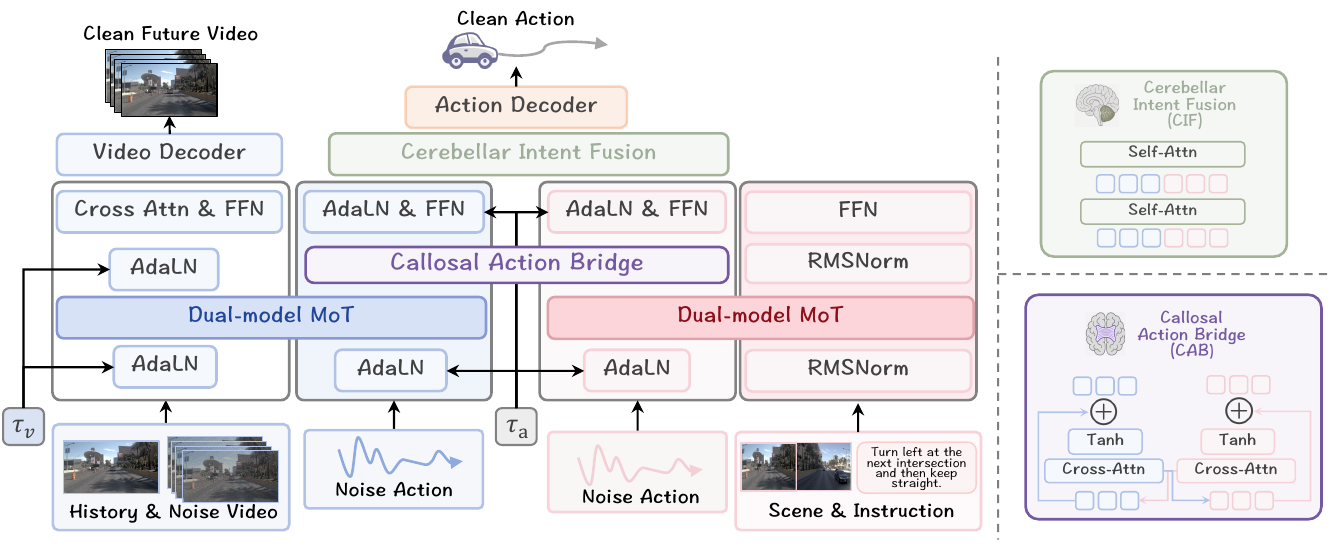}
    \caption{
    \textbf{Overview of the proposed semantic-predictive action architecture.}
    The VLA pathway distills scene semantics, route instructions, and rule-aware priors into semantic-grounded action tokens, while the WAM pathway distills future dynamics and physical priors into prediction-grounded action tokens.
    Instead of mixing raw VLM and VGM tokens in a shared attention space, CAB bridges the two action streams, and CIF fuses the refined action intents for trajectory decoding.
    }
    \label{fig:framework}
\end{figure*}

\paragraph{VLA Models for End-to-End Autonomous Driving.}
Vision-Language-Action (VLA) models have emerged in end-to-end autonomous driving, aiming to translate visual observations, route instructions, and traffic-scene semantics into executable trajectories~\citep{drivegpt4,lmdrive,drivevlm,drivelm}.
Early studies use large language or vision-language models for traffic-scene understanding, reasoning, and decision support, while recent methods move toward trajectory-level planning.
ORION~\citep{orion} bridges semantic reasoning and continuous action generation with a generative planner, ReCogDrive~\citep{recogdrive} couples a VLM with a diffusion planner, OpenDriveVLA~\citep{opendrivevla} builds an open VLA policy for driving actions, and AutoVLA~\citep{autovla} discretizes trajectories into action primitives for autoregressive policy learning.
These methods use VLM representations to guide action generation, but future scene evolution is not modeled as a planning signal. We therefore treat VLA as a semantic action pathway and coordinate it with a prediction-grounded WAM pathway.

\paragraph{World Models in Autonomous Driving.}
World models provide a route to planning by learning how driving scenes evolve over time~\citep{world4drive,mapworld}.
Early generative driving world models, such as GAIA-1~\citep{gaia1}, DriveDreamer~\citep{drivedreamer}, and ADriver-I~\citep{adriver_i}, show that video or vision-action generation can capture structured traffic evolution in real-world scenarios.
Recent methods connect world modeling with planning through future representation prediction~\citep{law}, joint image-action sequence modeling~\citep{drivinggpt,worldvla,univla}, controllable future generation~\citep{drivedreamer2}, and dense future supervision for policy learning~\citep{drivevla_w0}.
These approaches demonstrate the value of predictive modeling, especially in scenarios that require motion anticipation and physical consistency.
Most existing world-model methods emphasize future generation or use prediction as auxiliary supervision, whereas we use predictive representations as one action pathway and coordinate them with a separate semantic pathway.

\section{Method}
\label{sec:method}

In this section, we describe our coordination framework for VLA and WAM, as shown in Fig.~\ref{fig:framework}.
The method is trained in three stages.
First, the WAM branch learns prediction-grounded action representations from future scene dynamics (Sec.~\ref{sec:wam_branch}).
Second, the VLA branch learns semantic-grounded action representations from visual observations and language instructions (Sec.~\ref{sec:vla_branch}).
Finally, both branches are frozen, while CAB, CIF, and the final
action decoder are trained to coordinate the two action streams and
generate the final trajectory (Sec.~\ref{sec:joint}).

\begin{figure*}[!t]
    \centering
    \includegraphics[width=\textwidth,height=0.32\textheight,keepaspectratio]{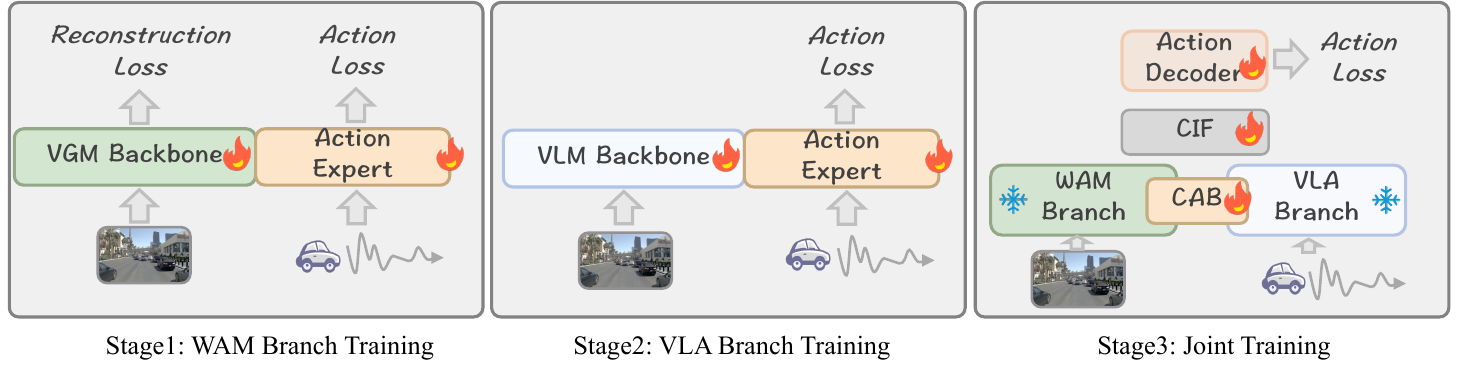}
    \caption{\textbf{Three-stage training pipeline.}
    Stage 1 trains the WAM branch with video and action rectified-flow objectives, enabling the action expert to learn prediction-grounded representations from future scene modeling.
    Stage 2 trains the VLA branch with visual and language inputs, converting VLM semantic cues into action representations.
    Stage 3 freezes both branches and optimizes CAB, CIF, and the action decoder for joint trajectory generation.
    }
    \label{fig:training_pipeline}
\end{figure*}

\subsection{WAM Branch}
\label{sec:wam_branch}

The WAM branch learns prediction-grounded action representations from future scene prediction.
Given the current observation, a video generative backbone predicts future visual latents, while a rectified-flow action expert generates the ego trajectory.
We perturb the video latent $x^v$ and the action trajectory $x^a$ with independent rectified-flow timesteps.
This decoupled schedule allows the video stream to terminate early after forming predictive context, while the action stream continues denoising to generate the trajectory.

\paragraph{Architecture.}

The WAM stream is shown on the left side of the main architecture in Fig.~\ref{fig:framework}.
We adopt Wan2.2-TI2V-5B~\citep{wan22} as the video backbone and attach a lightweight action expert.
The video backbone performs denoising over future video latents and produces visual tokens $V$ that capture scene dynamics, while the action expert performs trajectory denoising and produces action tokens $A_{\mathrm{pred}}$.
Dual-MoT modules couple the two streams through shared self-attention, enabling visual dynamics and action trajectories to interact, while modality-specific feed-forward networks preserve their distinct modeling capacities.

The branch predicts video and action vector fields:
\begin{equation}
    \hat{u}^{v},\, \hat{u}^{a}_{\mathrm{pred}}
    =
    F_{\mathrm{WAM}}\!\left(x^v_{t_v},\, x^a_{t_a},\, t_v,\, t_a,\, c_{\mathrm{obs}}\right),
    \label{eq:wam_forward}
\end{equation}
where $F_{\mathrm{WAM}}$ denotes the WAM stream with Dual-MoT interaction.
Here, $c_{\mathrm{obs}}$ is the conditioning feature from the current observation, $\hat{u}^{v}$ is the predicted vector field, and $\hat{u}^{a}_{\mathrm{pred}}$ is the predicted action vector field.

\paragraph{Rectified-flow training.}
We follow Flow Matching~\citep{lipman2023flow} and Rectified Flow~\citep{liu2023rectified}, which define a linear path between clean data $x_0$ and Gaussian noise $\epsilon$:
\begin{equation}
    x_t = (1 - t)\, x_0 + t\, \epsilon, \quad \epsilon \sim \mathcal{N}(0, I).
    \label{eq:fm_path}
\end{equation}
For the WAM branch, this path is applied to both future video latents and action trajectories:
\begin{equation}
    x^v_{t_v} = (1 - t_v)x^v_0 + t_v \epsilon^v,
    \qquad
    x^a_{t_a} = (1 - t_a)x^a_0 + t_a \epsilon^a .
    \label{eq:wam_noising}
\end{equation}
The corresponding velocity targets are
\begin{equation}
    u^v = \epsilon^v - x^v_0,
    \qquad
    u^a = \epsilon^a - x^a_0 .
    \label{eq:wam_velocity_target}
\end{equation}
We supervise the predicted video and action velocity fields with
\begin{equation}
\begin{aligned}
    \mathcal{L}_{\mathrm{vid}}
    &= \mathbb{E}\| \hat{u}^{v} - u^v \|_2^2, \\
    \mathcal{L}_{\mathrm{pred}}^{\mathrm{a}}
    &= \mathbb{E}\| \hat{u}^{a}_{\mathrm{pred}} - u^a \|_2^2 .
\end{aligned}
\label{eq:wam_losses}
\end{equation}
The total WAM loss is
\begin{equation}
    \mathcal{L}_{\mathrm{WAM}}
    =
    \mathcal{L}_{\mathrm{vid}}
    +
    \lambda_{\mathrm{pred}}^{\mathrm{a}}
    \mathcal{L}_{\mathrm{pred}}^{\mathrm{a}} .
    \label{eq:wam_total}
\end{equation}

\subsection{VLA Branch}
\label{sec:vla_branch}

The VLA branch uses a VLM backbone to extract semantics from visual observations and language instructions.
It complements the WAM branch through scene-level understanding and route-conditioned intent rather than future visual prediction.
A rectified-flow action expert converts VLM features into semantic-grounded action representations.

\paragraph{Architecture.}
The VLA stream is shown on the right side of the main architecture in Fig.~\ref{fig:framework}.
We adopt Qwen3-VL-4B~\citep{qwen3vl} as the VLM backbone and equip it with a lightweight action expert for trajectory modeling.
The VLM encodes multi-view images and driving instructions into semantic tokens $U$, and ego history into state tokens $E$.
The action expert processes the noisy trajectory $x^a_{t_a}$ into action tokens $A_{\mathrm{sem}}$.
Dual-MoT modules couple semantic, state, and action tokens through shared self-attention to guide action denoising.

The VLA branch predicts the action vector field:
\begin{equation}
    \hat{u}^{a}_{\mathrm{sem}}
    =
    F_{\mathrm{VLA}}\!\left(U,\, E,\, x^a_{t_a},\, t_a\right),
    \label{eq:vla_forward}
\end{equation}
where $F_{\mathrm{VLA}}$ denotes the Dual-MoT VLA stream, and $\hat{u}^{a}_{\mathrm{sem}}$ is the semantic-grounded action vector field.

\paragraph{Rectified-flow training.}
The VLA branch follows the action noising path defined in Eq.~\eqref{eq:wam_noising}.
Given the target velocity $u^a = \epsilon^a - x^a_0$, its training loss is
\begin{equation}
    \mathcal{L}_{\mathrm{sem}}^{\mathrm{a}}
    =
    \mathbb{E}\| \hat{u}^{a}_{\mathrm{sem}} - u^a \|_2^2 .
    \label{eq:vla_loss}
\end{equation}

\subsection{Joint Training with CAB and CIF}
\label{sec:joint}

\paragraph{Setup.}
As shown on the right side of Fig.~\ref{fig:training_pipeline}, the joint stage couples the pretrained WAM and VLA branches.
Both branches are frozen, while only CAB, CIF, and the final action decoder are optimized.
This preserves their pretrained modeling capabilities and focuses learning on cross-stream coordination.
Both action experts receive the same noisy trajectory and action timestep:
\begin{equation}
    x^a_{t_a} = (1 - t_a)\, x^a_0 + t_a\, \epsilon^a,
    \quad \epsilon^a \sim \mathcal{N}(0, I).
    \label{eq:joint_path}
\end{equation}
This places the action tokens from both streams at the same noise level, while the WAM video branch retains its own timestep $t_v$ to provide predictive context.

\paragraph{Callosal Action Bridge (CAB)}
Inspired by the corpus callosum connecting specialized hemispheres, CAB enables bidirectional interaction between prediction-grounded action tokens $A_{\mathrm{pred}}$ and semantic-grounded action tokens $A_{\mathrm{sem}}$.
Unlike token-level fusion, CAB avoids mixing raw VLM and video tokens within a shared attention pool.
At layer $l$, CAB computes bidirectional cross-stream messages:
\begin{equation}
\begin{aligned}
    M_{\mathrm{pred}\leftarrow\mathrm{sem}}^l
    &=
    \Psi_{\mathrm{cab}}^l
    \left(A_{\mathrm{pred}}^l, A_{\mathrm{sem}}^l\right), \\
    M_{\mathrm{sem}\leftarrow\mathrm{pred}}^l
    &=
    \Psi_{\mathrm{cab}}^l
    \left(A_{\mathrm{sem}}^l, A_{\mathrm{pred}}^l\right).
\end{aligned}
\label{eq:cab_messages}
\end{equation}
where $\Psi_{\mathrm{cab}}^l(X,Y)$ denotes cross-attention using $X$ as queries and $Y$ as keys and values.
The two action streams are subsequently updated through gated residual injection:
\begin{equation}
\begin{aligned}
    \tilde{A}_{\mathrm{pred}}^l
    &=
    A_{\mathrm{pred}}^l
    +
    \alpha_{\mathrm{pred}}^l
    M_{\mathrm{pred}\leftarrow\mathrm{sem}}^l, \\
    \tilde{A}_{\mathrm{sem}}^l
    &=
    A_{\mathrm{sem}}^l
    +
    \alpha_{\mathrm{sem}}^l
    M_{\mathrm{sem}\leftarrow\mathrm{pred}}^l .
\end{aligned}
\label{eq:cab_update}
\end{equation}
with learnable residual gates
$\alpha_{\mathrm{pred}}^l=\tanh(g_{\mathrm{pred}}^l)$
and
$\alpha_{\mathrm{sem}}^l=\tanh(g_{\mathrm{sem}}^l)$.
Following~\citep{alayrac2022flamingo,zhang2023llamaadapter}, the gates are zero-initialized, preserving pretrained action streams initially while learning cross-stream updates during joint training.

\paragraph{Cerebellar Intent Fusion (CIF)}
Inspired by the cerebellum’s role in motor coordination, CIF integrates the refined action streams into a unified representation.
It concatenates both streams, processes them with a lightweight Transformer module, and averages the resulting outputs:
\begin{equation}
\begin{gathered}
    Z_{\mathrm{pred}}, Z_{\mathrm{sem}}
    =
    \mathrm{CIF}\!\left(
    \tilde{A}_{\mathrm{pred}}^L,
    \tilde{A}_{\mathrm{sem}}^L
    \right), \\
    Z =
    \mathcal{M}(Z_{\mathrm{pred}}, Z_{\mathrm{sem}}).
\end{gathered}
\label{eq:cif}
\end{equation}
where $\mathcal{M}$ denotes element-wise averaging.
The fused representation is decoded into the action velocity:
$\hat{u}^a_{\mathrm{fuse}} = D_{\mathrm{fuse}}(Z, t_a)$.
Joint training supervises only the fused prediction:
\begin{equation}
    \mathcal{L}_{\mathrm{fuse}}
    =
    \mathbb{E}\| \hat{u}^a_{\mathrm{fuse}} - u^a \|_2^2,
    \quad
    u^a = \epsilon^a - x^a_0 .
    \label{eq:cif_loss}
\end{equation}

\paragraph{Inference.}
At inference, both action experts start from the same noise trajectory and follow identical timesteps.
CAB coordinates their intermediate representations, after which CIF fuses the two streams for final trajectory decoding.

\section{Experiments}
\label{sec:exp}

\subsection{Benchmark and Datasets}
\label{sec:exp_benchmark}

We evaluate planning performance on NAVSIM v1~\citep{navsim} and NAVSIM v2~\citep{navsim_v2}.
NAVSIM is built upon OpenScene~\citep{openscene}, a reprocessed version of nuPlan~\citep{nuplan}, and consists of real-world driving logs.
At each frame, the model predicts a $4$-second trajectory at $2 \mathrm{Hz}$, yielding $8$ waypoints.
The predicted trajectory is evaluated in a short-horizon, non-reactive simulation.
Unlike open-loop displacement metrics, this protocol additionally evaluates safety, driving progress, and rule compliance.

NAVSIM v1 reports the Predictive Driver Model Score (PDMS),
which aggregates No at-fault Collision (NC), Drivable Area Compliance (DAC),
Time-To-Collision (TTC), Comfort (C), and Ego Progress (EP).
NC and DAC serve as multiplicative safety penalties, while TTC and EP measure temporal risk and driving efficiency, respectively, and C evaluates ride comfort:
\begin{equation*}
\mathrm{PDMS} =
\mathrm{NC} \times \mathrm{DAC} \times
\frac{
5\,\mathrm{EP} + 5\,\mathrm{TTC} + 2\,\mathrm{C}
}{12}.
\end{equation*}

NAVSIM v2 extends this metric with two additional penalty multipliers, Driving Direction Compliance (DDC) and Traffic Light Compliance (TLC), and three weighted subscores, Lane Keeping (LK), History Comfort (HC), and Extended Comfort (EC).
The Extended PDMS (EPDMS) is defined as
\begin{equation*}
\mathrm{EPDMS}
=
\left(
\prod_{m \in \mathcal{M}_{\mathrm{pen}}}
s_m
\right)
\left(
\frac{
\sum_{m \in \mathcal{M}_{\mathrm{avg}}} w_m s_m
}{
\sum_{m \in \mathcal{M}_{\mathrm{avg}}} w_m
}
\right),
\label{eq:epdms}
\end{equation*}
Here, $\mathcal{M}_{\mathrm{pen}}=\{\mathrm{NC},\mathrm{DAC},\mathrm{DDC},\mathrm{TLC}\}$ and $\mathcal{M}_{\mathrm{avg}}=\{\mathrm{TTC},\mathrm{EP},\mathrm{HC},\mathrm{LK},\mathrm{EC}\}$.
The weights are $w_{\mathrm{TTC}}=w_{\mathrm{EP}}=5$ and $w_{\mathrm{HC}}=w_{\mathrm{LK}}=w_{\mathrm{EC}}=2$.

\subsection{Implementation Details}
\label{sec:exp_impl}

Each of the three stages is trained for 100K steps on 8 NVIDIA H20
GPUs with a per-GPU batch size of 6.
We use AdamW with a cosine learning-rate schedule, 200 warmup steps,
and a peak learning rate of $5\times10^{-5}$.
Training uses bf16 mixed precision, with checkpoints saved every 3K
steps.
At inference, we use 3-step rectified-flow sampling for the action
streams.

\subsection{Main Results}
\label{sec:exp_main}

\begin{table*}[t]
\centering
\small
\setlength{\tabcolsep}{4.5pt}
\caption{\textbf{Planning performance comparison on NAVSIM v1.}
Best results use \textbf{bold} and second-best results are \underline{underlined}.}
\label{tab:navsim_v1}
\begin{tabular}{l c c c c c c c c >{\columncolor{gray!20}}c}
\toprule
\textbf{Method} & \textbf{Ref.} & \textbf{Image} & \textbf{Lidar} & \textbf{NC}$\uparrow$ & \textbf{DAC}$\uparrow$ & \textbf{TTC}$\uparrow$ & \textbf{C}$\uparrow$ & \textbf{EP}$\uparrow$ & \textbf{PDMS}$\uparrow$ \\
\midrule
Human & -- & -- & -- & 100.0 & 100.0 & 100.0 & 99.9 & 87.5 & 94.8 \\
\midrule
\multicolumn{10}{l}{\emph{Traditional End-to-End Methods}} \\
\quad TransFuser~\citep{transfuser}       & TPAMI'23 & \checkmark & \checkmark & 97.7 & 92.8 & 92.8 & 100.0 & 79.2 & 84.0 \\
\quad UniAD~\citep{uniad}            & CVPR'23  & \checkmark &            & 97.8 & 91.9 & 92.9 & 100.0 & 78.8 & 83.4 \\
\quad PARA-Drive~\citep{paradrive}       & CVPR'24  & \checkmark &            & 97.9 & 92.4 & 93.0 & 99.8  & 79.3 & 84.0 \\
\quad DiffusionDrive~\citep{diffusiondrive}   & CVPR'25  & \checkmark & \checkmark & 98.2 & 96.2 & 94.7 & 100.0 & \underline{82.2} & 88.1 \\
\midrule
\multicolumn{10}{l}{\emph{Vision-Language-Action Methods}} \\
\quad ReCogDrive~\citep{recogdrive}   & ICLR'26    & \checkmark &            & 98.1 & 94.7 & 94.2 & 100.0 & 80.9 & 86.5 \\
\quad DynVLA~\citep{dynvla}             & ICML'26    & \checkmark &            & 98.6 & 95.3 & 95.5 & 100.0 & 80.6 & 87.2 \\
\quad AutoVLA~\citep{autovla}           & NeurIPS'25 & \checkmark &            & 98.4 & 95.6 & \textbf{98.0} & 99.9 & 81.9 & \underline{89.1} \\
\quad DriveVLA-W0~\citep{drivevla_w0}   & ICLR'26    & \checkmark &            & 98.4 & 95.3 & 95.2 & 100.0 & 80.9 & 87.2 \\
\midrule
\multicolumn{10}{l}{\emph{World-Model-Based Methods}} \\
\quad DrivingGPT~\citep{drivinggpt}     & ICCV'25 & \checkmark &            & \underline{98.9} & 90.7 & 94.9 & 95.6  & 79.7 & 82.4 \\
\quad LAW~\citep{law}                   & ICLR'25 & \checkmark &            & 96.4 & 95.4 & 88.7 & 99.9  & 81.7 & 84.6 \\
\quad Epona~\citep{epona}                             & ICCV'25 & \checkmark &            & 97.9 & 95.1 & 93.8 & 99.9  & 80.4 & 86.2 \\
\quad WoTE~\citep{wote}                              & ICCV'25 & \checkmark & \checkmark & 98.5 & 96.8 & 94.9 & 99.9  & 81.9 & 88.3 \\
\quad DriveLaW~\citep{drivelaw}                          & CVPR'26 & \checkmark &            & \textbf{99.0} & \underline{97.1} & \underline{96.7} & 100.0 & 81.3 & \underline{89.1} \\
\midrule
\textbf{BrainWAM (Ours)}                     & --      & \checkmark &            & 98.1 & \textbf{97.5} & 94.9 & 100.0 & \textbf{83.8} & \textbf{89.5} \\
\bottomrule
\end{tabular}
\end{table*}

\begin{table*}[t]
\centering
\small
\setlength{\tabcolsep}{4.5pt}
\caption{\textbf{Planning performance comparison on NAVSIM v2.} The benchmark evaluates driving performance
under additional rule-based and comfort-related metrics. The best results are highlighted in \textbf{bold}.}
\label{tab:navsim_v2}
\begin{tabular}{l c c c c c c c c c >{\columncolor{gray!20}}c}
\toprule
\textbf{Method} & \textbf{NC}$\uparrow$ & \textbf{DAC}$\uparrow$ & \textbf{DDC}$\uparrow$ & \textbf{TLC}$\uparrow$ & \textbf{EP}$\uparrow$ & \textbf{TTC}$\uparrow$ & \textbf{LK}$\uparrow$ & \textbf{HC}$\uparrow$ & \textbf{EC}$\uparrow$ & \textbf{EPDMS}$\uparrow$ \\
\midrule
\multicolumn{11}{l}{\emph{Traditional End-to-End Methods}} \\
\quad TransFuser~\citep{transfuser}     & 96.9 & 89.9 & 97.8 & 99.7 & 87.1 & 95.4 & 92.7 & 98.3 & 87.2 & 76.7 \\
\quad HydraMDP++~\citep{hydramdpp}     & 97.2 & 97.5 & 99.4 & 99.6 & 83.1 & 96.5 & 94.4 & 98.2 & 70.9 & 81.4 \\
\quad DriveSuprim~\citep{drivesuprim}    & 97.5 & 96.5 & 99.4 & 99.6 & \textbf{88.4} & 96.6 & 95.5 & 98.3 & 77.0 & 83.1 \\
\quad ARTEMIS~\citep{artemis}        & 98.3 & 95.1 & 98.6 & 99.8 & 81.5 & 97.4 & 96.5 & 98.3 & \textbf{89.1} & 83.1 \\
\midrule
\multicolumn{11}{l}{\emph{Vision-Language-Action Methods}} \\
\quad DriveVLA-W0~\citep{drivevla_w0}  & \textbf{98.5} & \textbf{99.1} & 98.0 & 99.7 & 86.4 & \textbf{98.1} & 93.2 & 97.9 & 58.9 & 86.1 \\
\midrule
\multicolumn{11}{l}{\emph{World-Model-Based Methods}} \\
\quad DriveDreamer-Policy~\citep{drivedreamer_policy} & 98.4 & 97.1 & 99.5 & \textbf{99.9} & 87.9 & 97.7 & \textbf{97.6} & 98.3 & 79.4 & 88.7 \\
\midrule
\textbf{BrainWAM (Ours)}  & 98.1 & 97.5 & \textbf{99.6} & \textbf{99.9} & 88.2 & 97.4 & \textbf{97.6} & \textbf{98.4} & 85.8 & \textbf{89.6} \\
\bottomrule
\end{tabular}
\end{table*}

\paragraph{NAVSIM v1 results.}
As shown in Table~\ref{tab:navsim_v1}, BrainWAM achieves a PDMS of $89.5$, outperforming both VLA-based and world-model-based baselines.
The gains are most pronounced in DAC and EP, indicating improved drivable-area compliance and driving progress, while maintaining competitive NC, TTC, and comfort scores.

\paragraph{NAVSIM v2 results.}
Table~\ref{tab:navsim_v2} shows that BrainWAM achieves state-of-the-art performance on NAVSIM v2, with an EPDMS of $89.6$.
The improvements are primarily driven by EP and EC, whereas several rule-compliance metrics are already near saturation.
These results demonstrate that the proposed coordination remains effective under the more comprehensive NAVSIM v2 evaluation protocol.

\subsection{Further Analysis and Ablation Studies}
\label{sec:exp_ablation}

\begin{table}[t]
\centering
\small
\setlength{\tabcolsep}{4pt}
\caption{\textbf{Ablation of branch and coordination strategies on NAVSIM v1.}}
\label{tab:ablation_pathway}
\begin{tabular}{l c c c c c >{\columncolor{gray!20}}c}
\toprule
\textbf{Method}   & \textbf{NC}$\uparrow$   & \textbf{DAC}$\uparrow$  & \textbf{TTC}$\uparrow$  & \textbf{C}$\uparrow$ & \textbf{EP}$\uparrow$   & \textbf{PDMS}$\uparrow$ \\
\midrule
VLA-only & 97.7 & 94.9 & 93.3 & 100.0 & 80.7 & 86.1 \\
WAM-only & 98.0 & 96.4 & 94.4 & 100.0 & 82.6 & 88.1 \\
Tri-MoT  & \textbf{98.3} & 96.2 & 94.7 & 100.0 & 81.7 & 87.8 \\
\textbf{BrainWAM}     & 98.1 & \textbf{97.5} & \textbf{94.9} & 100.0 & \textbf{83.8} & \textbf{89.5} \\
\bottomrule
\end{tabular}
\end{table}

\paragraph{Branch complementarity.}
Table~\ref{tab:ablation_pathway} compares the full model with its single-branch variants.
WAM-only achieves $88.1$ PDMS and substantially outperforms VLA-only, demonstrating the strong planning prior provided by predictive modeling on NAVSIM.
BrainWAM further improves PDMS to $89.5$, exceeding both single-branch variants.
This improvement suggests that semantic and predictive action representations provide complementary information under action-level coordination.

\paragraph{Action-level coordination vs.\ token-level fusion.}
As shown in Table~\ref{tab:ablation_pathway}, Tri-MoT achieves $87.8$ PDMS, underperforming the WAM-only variant.
This indicates that directly mixing VLM and video tokens in a shared attention space does not effectively transfer semantic knowledge to planning.
Fig.~\ref{fig:trimot_attn} further reveals imbalanced cross-modal attention in Tri-MoT, motivating coordination at the action level.
By keeping raw modality tokens separate and interacting only through action representations, BrainWAM improves PDMS to $89.5$.
Because both methods use identical backbones and comparable parameter counts, the gain is attributable to the coordination mechanism rather than increased model capacity.

\begin{table}[t]
\centering
\small
\setlength{\tabcolsep}{4pt}
\caption{\textbf{Ablation study on NAVSIM v1, analyzing the effect of CAB and CIF.} The best results are highlighted in bold.}
\label{tab:ablation_cabcif}
\begin{tabular}{c c c c c c c >{\columncolor{gray!20}}c}
\toprule
\textbf{CAB} & \textbf{CIF} & \textbf{NC}$\uparrow$ & \textbf{DAC}$\uparrow$ & \textbf{TTC}$\uparrow$ & \textbf{C}$\uparrow$ & \textbf{EP}$\uparrow$ & \textbf{PDMS}$\uparrow$ \\
\midrule
\checkmark &            & 98.1 & 96.8 & 94.8 & 100.0 & 83.0 & 88.7 \\
           & \checkmark & 98.1 & 96.7 & 94.7 & 100.0 & 82.9 & 88.5 \\
\checkmark & \checkmark & 98.1 & \textbf{97.5} & \textbf{94.9} & 100.0 & \textbf{83.8} & \textbf{89.5} \\
\bottomrule
\end{tabular}
\end{table}

\paragraph{Effectiveness of CAB and CIF}
Table~\ref{tab:ablation_cabcif} evaluates the individual contributions of CAB and CIF.
Using CAB or CIF alone yields $88.7$ and $88.5$ PDMS, respectively, whereas combining them increases PDMS to $89.5$.
The improvement is concentrated in DAC and EP, while NC and TTC remain stable.
These results suggest that CAB facilitates intermediate interaction between the two action streams, while CIF consolidates their final representations.

\begin{table}[t]
    \centering
    \small
    \caption{\textbf{Trade-off between video denoising steps, planning performance, and inference latency.}
    All inference latencies are measured on a single NVIDIA H20 GPU.
    The best results are highlighted in bold.}
    \label{tab:ablation_denoise}
    \begin{tabular}{c c >{\columncolor{gray!20}}c >{\columncolor{gray!20}}c}
    \toprule
    \textbf{Video denoise steps} & \textbf{Latency}$\downarrow$ & \textbf{PDMS}$\uparrow$ & \textbf{EPDMS}$\uparrow$ \\
    \midrule
    0 & \textbf{382\,ms} & 79.3 & 75.8 \\
    1 & 475\,ms & 89.3 & 89.4 \\
    2 & 565\,ms & \textbf{89.5} & \textbf{89.6} \\
    3 & 644\,ms & 89.4 & \textbf{89.6} \\
    \bottomrule
    \end{tabular}
    \end{table}

\begin{figure*}[!t]
    \centering
    \includegraphics[width=0.8\textwidth]{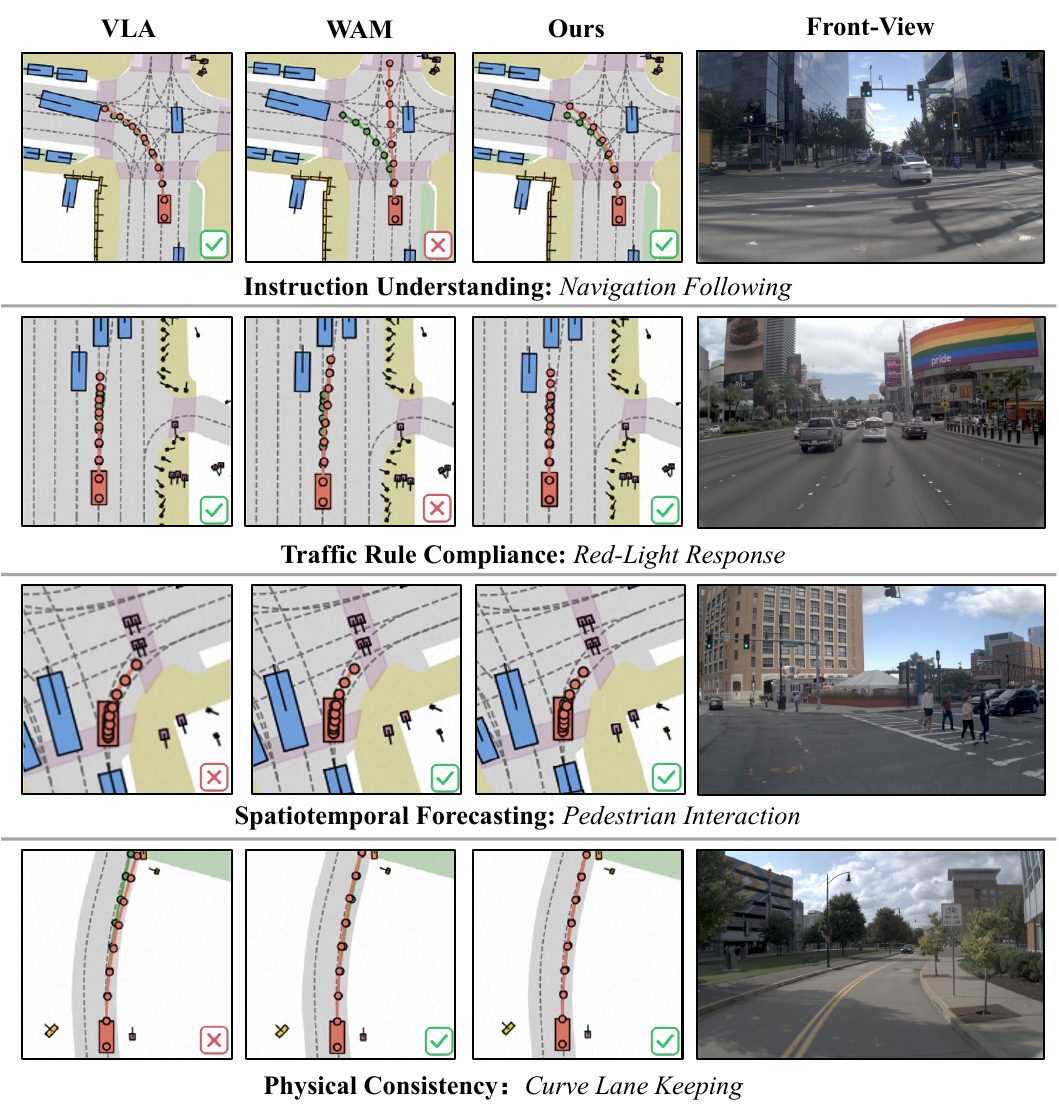}
    \caption{\textbf{Qualitative comparison of VLA-only, WAM-only, and Ours
    in representative scenarios.}
    Ours produces robust trajectories under semantic constraints
    and dynamic interactions by combining the two pathways at the action-token level.}
    \label{fig:qualitative}
\end{figure*}

\paragraph{Asynchronous video denoising.}
Table~\ref{tab:ablation_denoise} examines the number of video denoising steps used at inference.
Because the video and action streams follow independent rectified-flow timesteps, the video branch uses a truncated schedule and caches its features for subsequent action denoising.
With no video denoising, the model loses predictive context and drops to $79.3$ PDMS and $75.8$ EPDMS, confirming that video dynamics are essential to planning.
A single video step restores PDMS to $89.3$, after which performance remains between $89.3$ and $89.5$ as the number of steps increases to $3$, while latency rises from $475$ ms to $644$ ms.
Thus, one early video step provides most of the useful predictive context, offering a favorable trade-off between accuracy and efficiency.

\subsection{Qualitative Analysis}
\label{sec:exp_qualitative}

Fig.~\ref{fig:qualitative} compares VLA-only, WAM-only, and BrainWAM across representative scenarios.
These cases include semantic-grounding challenges, such as navigation following and brake-light understanding, as well as future-modeling challenges, such as interactive negotiation and trajectory feasibility.

In navigation following, the planner follows the route instruction rather than choosing a locally plausible but incorrect branch.
In red-light understanding, the planner jointly interprets the braking signal of the lead vehicle and the red traffic light to avoid a rear-end collision.
VLA-only handles these cases better than WAM-only, demonstrating its advantage in instruction grounding and semantic scene understanding.

Interactive negotiation involves coupled behaviors among the ego vehicle, pedestrians, and surrounding agents, where feasible planning depends on anticipating how the scene may evolve.
Trajectory feasibility is particularly challenging on curved roads, where planning solely from the current observation may produce inaccurate future motion.
WAM-only performs better in these cases, benefiting from jointly modeling future scene evolution and ego actions.

BrainWAM handles all four cases by coordinating semantic-grounded and prediction-grounded action representations.
This coordination reduces the failure modes observed in the single-branch variants.

\section{Conclusion}
\label{sec:conclusion}
In this work, we study how to effectively combine VLA-based semantic reasoning and WAM-based predictive world modeling for end-to-end autonomous driving. We first reveal that naive shared-token fusion suffers from an \emph{attention-allocation mismatch}: action tokens attend disproportionately to semantic tokens, which weakens the predictive signals provided by the world model, resulting in suboptimal planning. Motivated by this, we propose BrainWAM, which allows the two branches to first produce \emph{semantic-grounded} and \emph{prediction-grounded} action representations, and then coordinate them through structured interaction in a unified action space, while preserving the complementary specialization of semantic and predictive pathways. BrainWAM achieves state-of-the-art performance on both NAVSIM v1 and v2, demonstrating its effectiveness and potential for autonomous driving systems. We further analyze the limitations and future work in the supplementary material.

\bibliographystyle{ACM-Reference-Format}
\bibliography{references}

\appendix

\twocolumn[
\begin{center}
    {\LARGE\bfseries Appendix\par}
\end{center}
\vspace{0.8em}
]

\section{Modality Imbalance in Tri-MoT}

The attention imbalance observed in Tri-MoT is closely related to modality competition in multimodal learning. When heterogeneous modalities are jointly optimized in a shared representation space, the model tends to rely on the modality that offers more stable and easily optimizable signals, and this modality dominates the joint training~\citep{gradient_blending,modality_competition}. Such dominance can suppress the complementary modality and even make joint training underperform the best single-modality model~\citep{gradient_blending,unimodal_feature_learning}.

This mechanism explains the behavior of Tri-MoT. The VLM tokens come from large-scale vision-language pretraining and encode compact semantic abstractions such as traffic rules, signals, and scene-level layout. They are clean and stable throughout training. The VGM tokens, in contrast, are produced by a rectified-flow denoising process that gradually refines representations from Gaussian noise, so their features are less stable, especially in early denoising stages. Following modality competition, the action tokens take the VLM tokens as the easier-to-learn modality and assign them higher attention, while the predictive dynamics carried by the VGM tokens are underused~\citep{ogm_ge,unimodal_feature_learning}.

Two observations in our experiments support this account and rule out simpler explanations. First, the low attention assigned to VGM tokens does not imply that
they are uninformative: when video denoising is disabled, PDMS drops
to 79.3, compared with 89.3--89.5 when one to three video denoising
steps are used (Table~\ref{tab:ablation_denoise}).
This confirms that the predictive context provided by the video
stream is essential for planning.
Second, adding VLM tokens to the shared attention space does not help: Tri-MoT reaches only $87.8$ PDMS and stays below the WAM-only model ($88.1$, Table~\ref{tab:ablation_pathway}), even though it has access to strictly more information. Together, these results indicate that the problem is not a lack of useful signal in either modality, but the competition that suppresses the denoising VGM stream once it shares one attention space with the clean VLM stream.

This analysis motivates our design. Instead of mixing raw VLM and VGM tokens in one attention space, BrainWAM lets each branch first form its own action representation and coordinates the two branches only at the action level, which avoids direct competition between the clean and the denoising modalities.

\section{Implementation and Analysis of CAB}
\label{app:cab}
Unless otherwise specified, the architectural ablations in
Tables~\ref{tab:ablation_cab_number}--\ref{tab:ablation_cif_depth}
use 10-step joint denoising for both the video and action streams
to provide a controlled comparison.
The main results instead use the asynchronous inference schedule
described in the main text, with 3-step action sampling and
truncated video denoising.

\subsection{Implementation Details}
\label{app:cab_implementation}

CAB operates on the two action-token streams, each containing
$L=8$ tokens with a hidden dimension of $1024$.
We insert CAB at Layers 9 and 18 of the two action experts.
Each CAB contains two parallel multi-head cross-attention modules:
one updates the prediction-grounded action tokens using the
semantic-grounded tokens as context, while the other performs the
reverse update.

Each cross-attention module uses $8$ heads with a head dimension of
$128$.
The query is obtained from the stream being updated, while the key
and value are obtained from the other stream.
Separate normalization layers are applied to the query and context
streams, and the query, key, value, and output projections do not
use bias.

The cross-attention output is injected through a gated residual:

\begin{equation*}
\widetilde{A}_{x}^{l}
=
A_{x}^{l}
+
\tanh(g_{x}^{l})
\odot
\operatorname{Attn}
\left(
A_{x}^{l},
A_{y}^{l}
\right)
\end{equation*}

where $x,y\in\{\mathrm{pred},\mathrm{sem}\}$ and $x\neq y$.
The gate $g_{x}^{l}\in\mathbb{R}^{1024}$ is initialized to zero,
such that CAB starts as an identity mapping and gradually learns
cross-stream residual updates during Stage 3.
The two CAB blocks contain approximately $16.8$M parameters in
total.

\subsection{Two CAB Blocks Are Sufficient}
\label{app:cab_number}

We evaluate the effect of using different numbers of CAB blocks.
The default configuration inserts two CABs at Layers 9 and 18.
For all variants, the video and action streams are jointly denoised
for 10 steps.
The results are reported in
Table~\ref{tab:ablation_cab_number}.

\begin{table}[t]
\centering
\small
\setlength{\tabcolsep}{4pt}
\caption{\textbf{Ablation on the number of CAB blocks on NAVSIM v1.}
Both the video and action streams are jointly denoised for 10 steps
in all configurations.}
\label{tab:ablation_cab_number}
\begin{tabular}{l c c c c c >{\columncolor{gray!20}}c}
\toprule
\textbf{\# CAB} &
\textbf{NC}$\uparrow$ &
\textbf{DAC}$\uparrow$ &
\textbf{TTC}$\uparrow$ &
\textbf{C}$\uparrow$ &
\textbf{EP}$\uparrow$ &
\textbf{PDMS}$\uparrow$ \\
\midrule
1  & 98.1 & 97.0 & 94.8 & 100.0 & 83.0 & 88.9 \\
2 & 98.3 & 97.4 & 95.0 & 100.0 & 83.5 & 89.3 \\
3  & 98.2 & 97.3 & 94.6 & 100.0 & 83.8 & 89.2 \\
5  & 98.2 & 97.4 & 94.8 & 100.0 & 83.7 & 89.3 \\
28 & 98.2 & 97.4 & 95.0 & 100.0 & 83.6 & 89.3 \\
\bottomrule
\end{tabular}
\end{table}

Using a single CAB yields 88.9 PDMS, showing that one interaction
layer is insufficient for fully coordinating the two action streams.
Increasing the number of CAB blocks to two improves PDMS to 89.3.
Further increasing the number to 3, 5, or 28 yields comparable
performance within a narrow range of 89.2--89.3 PDMS.

These results indicate that cross-stream communication largely
saturates after two CAB interactions.
We therefore use two CAB blocks at Layers 9 and 18, which match the
best performance of denser configurations with lower parameter and
computational overhead.

\section{Implementation and Analysis of CIF}
\label{app:cif}

\subsection{Implementation Details}
\label{app:cif_implementation}

CIF operates on two action streams, each containing $L=8$ tokens
with a hidden dimension of $1024$.
The two streams are first projected separately to a shared
$1024$-dimensional space, with a learnable source embedding added
to distinguish their origins.
The concatenated sequence is processed by a $2$-layer Transformer
with $8$ attention heads.
Each layer uses action-timestep-conditioned AdaLN modulation.
The timestep condition is obtained from a sinusoidal embedding
followed by an MLP.
CIF contains approximately $49.3$M parameters.

\subsection{Transformer-Based Fusion Performs Best}
\label{app:cif_fusion}

We compare three implementations of CIF: direct projection of the
concatenated tokens using an MLP, gated fusion, and the
Transformer-based fusion used in BrainWAM.
The video and action streams are jointly denoised for 10 steps in
all configurations.
The results are reported in
Table~\ref{tab:ablation_cif_fusion}.

\begin{table}[t]
\centering
\small
\setlength{\tabcolsep}{4pt}
\caption{\textbf{Ablation on CIF fusion strategies on NAVSIM v1.}
Both the video and action streams are jointly denoised for 10 steps
in all configurations.}
\label{tab:ablation_cif_fusion}
\begin{tabular}{l c c c c c >{\columncolor{gray!20}}c}
\toprule
\textbf{Fusion} &
\textbf{NC}$\uparrow$ &
\textbf{DAC}$\uparrow$ &
\textbf{TTC}$\uparrow$ &
\textbf{C}$\uparrow$ &
\textbf{EP}$\uparrow$ &
\textbf{PDMS}$\uparrow$ \\
\midrule
MLP                  & 97.9 & 96.9 & 94.1 & 100.0 & 83.8 & 88.8 \\
Gate                 & 98.0 & 97.2 & 94.4 & 100.0 & 83.9 & 89.1 \\
\textbf{Transformer} & 98.3 & 97.4 & 95.0 & 100.0 & 83.5 & \textbf{89.3} \\
\bottomrule
\end{tabular}
\end{table}

Direct MLP projection obtains 88.8 PDMS, while gated fusion improves
the result to 89.1.
The Transformer-based design achieves the best performance of
89.3 PDMS.
This comparison shows that token-level interaction is more effective
than direct projection or feature-wise gating for integrating the
two action streams.

\subsection{Two Transformer Layers Are Sufficient}
\label{app:cif_depth}

We further vary the number of Transformer layers in CIF.
All other architectural and inference settings remain unchanged,
and the video and action streams are jointly denoised for 10 steps.
The results are shown in Table~\ref{tab:ablation_cif_depth}.

Increasing the Transformer depth from one to two layers improves
PDMS from 89.0 to 89.3.
A third layer brings no further gain.
We therefore adopt two Transformer layers, which achieve the same
performance as the deeper variant with lower computational and
parameter overhead.

\begin{table}[htbp]
\centering
\small
\setlength{\tabcolsep}{4pt}
\caption{\textbf{Ablation on the Transformer depth of CIF on
NAVSIM v1.} Both the video and action streams are jointly denoised for 10 steps
in all configurations.
}
\label{tab:ablation_cif_depth}
\begin{tabular}{l c c c c c >{\columncolor{gray!20}}c}
\toprule
\textbf{\# Layers} &
\textbf{NC}$\uparrow$ &
\textbf{DAC}$\uparrow$ &
\textbf{TTC}$\uparrow$ &
\textbf{C}$\uparrow$ &
\textbf{EP}$\uparrow$ &
\textbf{PDMS}$\uparrow$ \\
\midrule
1          & 98.1 & 97.1 & 94.8 & 100.0 & 83.4 & 89.0 \\
2 & 98.3 & 97.4 & 95.0 & 100.0 & 83.5 & \textbf{89.3} \\
3          & 98.3 & 97.4 & 95.1 & 100.0 & 83.4 & \textbf{89.3} \\
\bottomrule
\end{tabular}
\end{table}

\section{Freezing the Pretrained Branches Stabilizes Stage-3 Optimization}
\label{app:stage3_update}

In Stage 3, we freeze the pretrained WAM and VLA branches and
optimize only CAB, CIF, and the final action decoder.
We compare this selective update strategy with end-to-end
fine-tuning of the entire model in
Table~\ref{tab:appendix_stage3}.

\begin{table}[t]
\centering
\small
\setlength{\tabcolsep}{6pt}
\caption{\textbf{Ablation on the Stage-3 update strategy on
NAVSIM v1.}
The selective strategy freezes the pretrained WAM and VLA branches
and updates only CAB, CIF, and the final action decoder.}
\label{tab:appendix_stage3}
\begin{tabular}{l >{\columncolor{gray!20}}c}
\toprule
\textbf{Stage-3 update strategy} &
\textbf{PDMS}$\uparrow$ \\
\midrule
Full-model fine-tuning
    & 88.8 \\
CAB, CIF, and action decoder only
    & \textbf{89.5} \\
\bottomrule
\end{tabular}
\end{table}

As shown in Table~\ref{tab:appendix_stage3}, full-model
fine-tuning obtains 88.8 PDMS, whereas selectively updating CAB,
CIF, and the action decoder improves PDMS to 89.5.
One reason is that the WAM and VLA branches exhibit different
convergence speeds during independent training.
The VLA-only branch reaches 86.1 PDMS after 54K steps, while the
WAM-only branch requires 81K steps to reach 88.1 PDMS.

When both branches are unfrozen and optimized jointly, their
different convergence speeds lead to unbalanced updates between the
two pathways.
This makes the representations received by CAB and CIF continuously
change at different rates, making stable coordination more difficult.
Moreover, end-to-end fine-tuning may disturb the complementary
representations acquired during branch-wise pretraining.

Freezing the two pretrained branches avoids this optimization
imbalance and provides stable inputs to CAB and CIF.
Stage 3 can therefore focus on coordinating and fusing the two
action representations rather than simultaneously adapting the two
large backbones.
This selective update strategy results in more stable optimization
and better planning performance.

\section{Additional Implementation Details}
\label{app:implementation}

All experiments are conducted on 8 NVIDIA H20 GPUs with a per-GPU
batch size of 6.
We use DeepSpeed ZeRO-2 and \texttt{bf16} mixed-precision training.
The model is optimized with AdamW using a peak learning rate of
$5\times10^{-5}$ and a weight decay of $0.01$.
The learning rate follows a cosine decay schedule with 200 warmup
steps.
Training runs for 100K optimization steps, with checkpoints saved
every 3K steps.

The three training stages use the same optimization configuration.
In Stages 1 and 2, the WAM and VLA branches are initialized from
their respective pretrained backbones and optimized independently.
In Stage 3, both pretrained branches are frozen, while CAB, CIF, and
the final action decoder are jointly optimized.

For inference, action generation uses 3-step rectified-flow sampling.
Under the asynchronous denoising schedule, the video branch is
stopped earlier than the action branch, and its intermediate features
are cached and reused by subsequent action denoising steps.
This avoids repeatedly evaluating the video backbone after its
denoising process has terminated.

\section{Additional Qualitative Comparisons}

\begin{figure*}[htbp]
    \centering
    \includegraphics[width=0.9\textwidth]{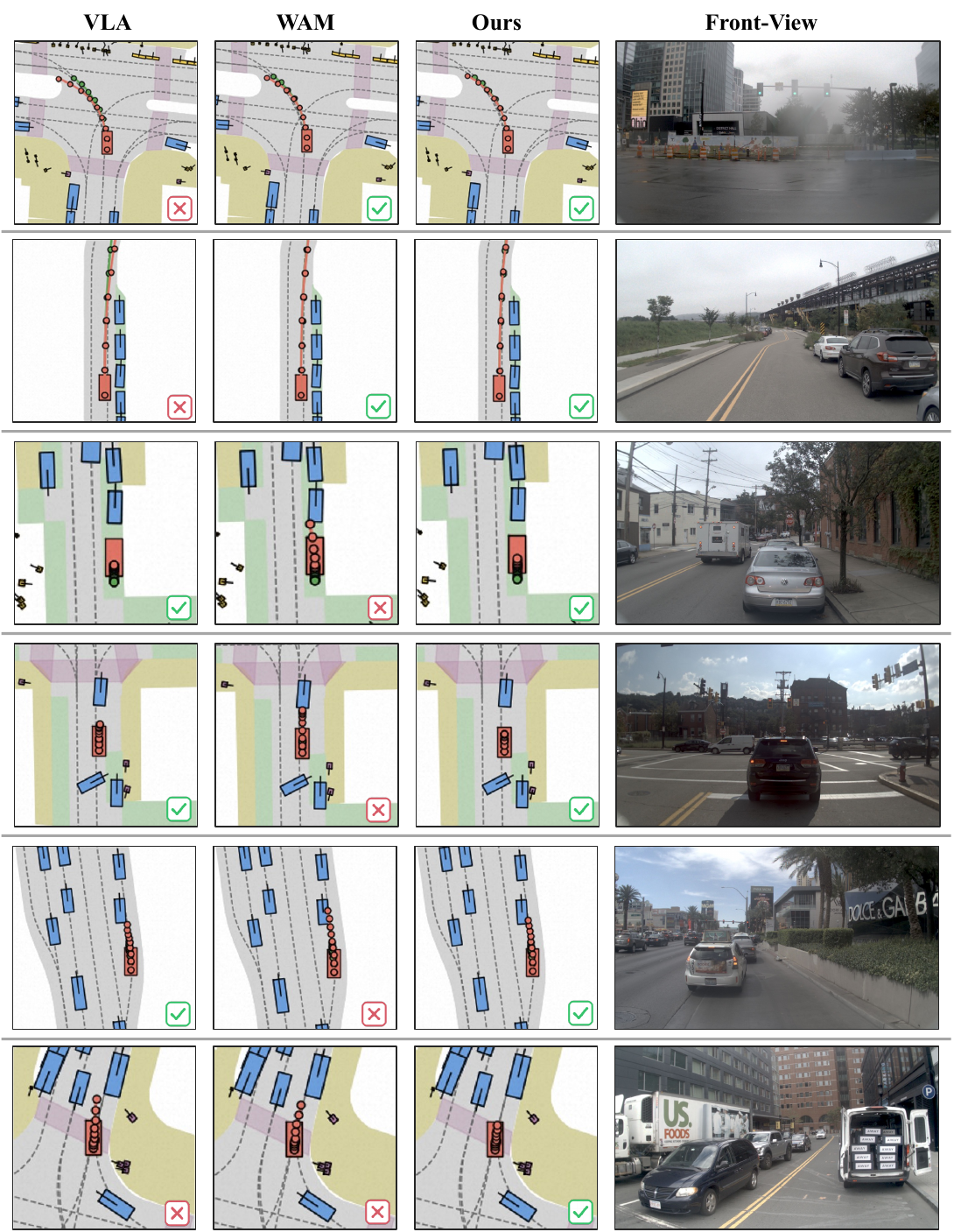}
    \caption{\textbf{Additional qualitative comparisons among VLA-only,
    WAM-only, and BrainWAM.}
    Each row presents the predicted trajectory in the BEV representation
    and the corresponding front-view image.
    The first two rows show cases where WAM-only succeeds while VLA-only
    fails, whereas Rows 3--5 show the opposite.
    In the last row, both single-branch models fail, while BrainWAM still
    produces a reasonable trajectory.
    These examples demonstrate the complementary strengths of semantic
    priors and predictive dynamics, as well as the effectiveness of their
    action-space coordination in BrainWAM.}
    \label{fig:appendix}
\end{figure*}
To further illustrate the complementary behaviors of the two
branches, we provide additional qualitative comparisons among
VLA-only, WAM-only, and BrainWAM in
Fig.~\ref{fig:appendix}.
The selected cases cover complex intersections, dense traffic, and
vehicle interactions under different road layouts.
VLA-only and WAM-only exhibit different failure patterns, whereas
BrainWAM generally produces more reliable trajectories by combining
semantic driving priors with predictive dynamics.

\section{Limitations}
\label{app:limitations}

BrainWAM jointly executes the WAM and VLA branches and retains a
generative video backbone during inference.
Consequently, its computational and memory costs remain higher than
those of a single-branch planner.
Although the asynchronous denoising schedule reduces inference
latency to $475$--$644$ ms, as reported in the main text,
this runtime does not yet satisfy the strict real-time requirements
of practical in-vehicle deployment.

Further efficiency improvements may require compressing or
distilling the video branch, reducing redundant computation between
the two pathways, and developing more aggressive feature-reuse or
early-exit strategies.
Therefore, improving deployment efficiency while preserving the
complementary semantic and predictive capabilities of BrainWAM
remains an important direction for future work.

\end{document}